\documentclass{article}

\usepackage{fullpage}
\usepackage[utf8]{inputenc} 
\usepackage[T1]{fontenc}    
\usepackage{hyperref}       
\usepackage{xcolor}       
\usepackage{url}            
\usepackage{booktabs}       
\usepackage{amsfonts}       
\usepackage{nicefrac}       
\usepackage{microtype}      

\usepackage{microtype}
\usepackage{graphicx}
\usepackage{subfigure}
\usepackage{booktabs} 
\usepackage{algorithm}
\usepackage[noend, compatible]{algpseudocode}
\usepackage{natbib}
\setcitestyle{authoryear,round,citesep={;},aysep={,},yysep={;}}
\renewcommand{\cite}[1]{\citep{#1}}

\usepackage{authblk}

\usepackage{amsmath,amsthm, mathtools, amssymb, bbm, varwidth}

\renewcommand{\algorithmiccomment}[1]{\bgroup\hfill$\triangleright$~#1\egroup}

\newtheorem{theorem}{Theorem}

\newtheorem{lemma}[theorem]{Lemma}

\def\Regret{\textnormal{Regret}}
\def\prev{\textnormal{prev}} 

\DeclareMathOperator*{\argmax}{arg\,max\,}

\DeclareMathOperator*{\expt}{\mathbb{E}}

\title{Regret Bounds for Discounted MDPs}

\author{Shuang Liu\thanks{s3liu@eng.ucsd.edu} }

\author{Hao Su\thanks{haosu@eng.ucsd.edu} }

\affil{
  University of California, San Diego
}
\begin{document}
\maketitle

\begin{abstract}
Reinforcement learning (RL) has traditionally been understood from an episodic perspective; the concept of non-episodic RL, where there is no restart and therefore no reliable recovery, remains elusive. A fundamental question in non-episodic RL is how to measure the performance of a learner and derive algorithms to maximize such performance. Conventional wisdom is to maximize the difference between the average reward received by the learner and the maximal long-term average reward. In this paper, we argue that if the total time budget is relatively limited compared to the complexity of the environment, such comparison may fail to reflect the finite-time optimality of the learner. We propose a family of measures, called $\gamma$-regret, which we believe to better capture the finite-time optimality. We give motivations and derive lower and upper bounds for such measures. {\color{red} A follow-up work~\cite{he2020nearly} has improved both our lower and upper bound, the gap is now closed at $\tilde{\Theta}\left(\frac{\sqrt{SAT}}{(1 - \gamma)^{\frac{1}{2}}}\right)$}.
\end{abstract}

\section{Introduction}
\label{sec: intro}
Reinforcement learning (RL) is concerned with how an algorithm should interact with a (partially) unknown Markov decision process (MDP) to maximize the cumulative reward. We distinguish between two types of RL: episodic RL, where the learner is reset to a starting distribution periodically; and non-episodic RL, where the learner strictly operates on the MDP without interruption. 
Theoretical analysis has shown that in the tabular setting, an episodic RL algorithm can be expected to perform almost as well as the optimal episodic non-stationary policy in the long run~\cite{azar2017minimax, zanette2019tighter, dann2019strategic}. 

In this work, we are interested in the metric of evaluating non-episodic RL algorithms. In literature, the average reward received by an RL algorithm has been compared with the optimal \emph{gain}~\cite{mahadevan1996average} a stationary policy can achieve from the starting state~\cite{jaksch2010near, bartlett2012regal, fruit2018efficient, ortner2020regret}. Intuitively, the optimal gain is the maximal average-reward a policy can achieve when it operates on the MDP for infinitely many steps. 

\begin{figure}
\centering
\includegraphics[width=0.6\linewidth]{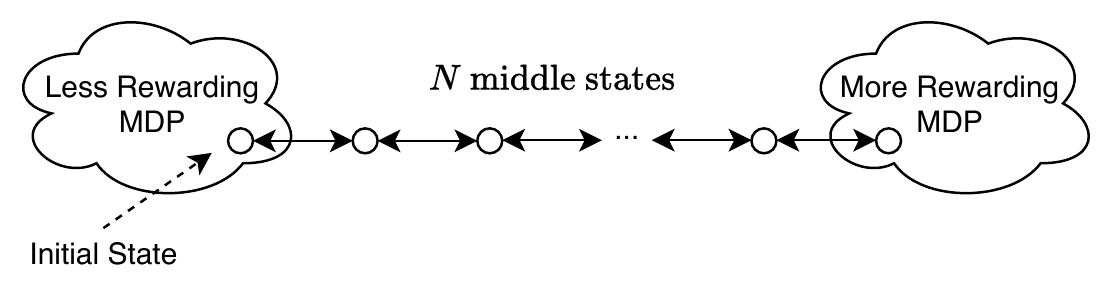}
\caption{An MDP formed by two sub-MDPs connected by $N$ middle states.}
\label{fig: ex}
\end{figure}

However, there exist certain nuisances in comparing a learning algorithm that interacts with the MDP for only a finite amount of steps (say, $T$ steps) with a policy that runs on the MDP for an infinite amount of steps. Previous work has rationalized this comparison by assuming the MDP has relatively short ``mixing time''. While the concept of mixing time is only formally defined for Markov chains, different authors have different interpretations of it in the case of MDPs. For example,~\citet{jaksch2010near} assumes the MDP has diameter $D \ll T$, and~\citet{bartlett2012regal, fruit2018efficient} assume the MDP weakly communicates and the optimal bias vector $h^*$ has bias-span $\textnormal{sp}(h^*) \ll T$; \citet{ortner2020regret} chooses to consider the maximal mixing time of the Markov chains induced by all the policies. To give a concrete example, consider the MDP in Figure~\ref{fig: ex}. The MDP is formed by two sub-MDPs connected by $N$ middle states. One sub-MDP is less rewarding, with reward range $[0, 1]$; the other one is more rewarding, with reward range $[2, 3]$. The learner starts from a state in the less rewarding MDP. It may choose to traverse between two sub-MDPs, but every middle state incurs a reward of $-1$. Obviously, given enough time (i.e., $T\gg N$), a reasonable learner should aim at arriving at the more rewarding MDP and stay there afterward. However, as long as $T \leq N$, it makes no sense for the learner to even leave the less rewarding MDP. 

An important observation that can be made from the above example is that, if the total time budget $T$ of a learner is short, it should be more myopic in order to earn more rewards. 
This observation motivates the research question: How to define a spectrum of measures that allow us to inspect the performance of an RL learner with a different time budget $T$? We thus propose a family of alternative optimality measures, called $\gamma$-regret, which we believe to better capture a learning algorithm's $T$-step optimality, and therefore can potentially be used as a guidance for deriving better algorithms. We will continue the discussion after formally defining $\gamma$-regret in the next section.

\section{Preliminaries}
\label{sec: prelim}
For a set $X$, denote by $\Delta(X)$ the set of probability distributions over $X$. A tabular MDP is defined by a finite state space $\mathcal{S} = \{1, 2, \cdots, S\}$, a finite action space $\mathcal{A} = \{1, 2, \cdots, A\}$, a transition function $M: \mathcal{S}\times\mathcal{A}\to\Delta(\mathcal{S})$, and a reward function $R:\mathcal{S}\times\mathcal{A}\to \Delta([0, 1])$. Unlike in the \emph{planning} setting, where $\mathcal{S}, \mathcal{A}, M, R$ are all known to the learner in advance, we consider the \emph{learning} setting where only $\mathcal{S}$ and $\mathcal{A}$ are known beforehand. 

An MDP can be interacted with either by calling \textsc{reset}, which returns an initial state $s_0$ and sets  $s_0$ as the current state; or by calling $\textsc{next}(a)$ if a current state $s$ is available, in which case a new state $s'\sim M(s, a)$ and a reward $r\sim R(s, a)$ are returned and $s'$ is set as the current state. 

We consider the \emph{non-episodic} setting, where the learner calls \textsc{reset} once at the very beginning, and then repeatedly calls \textsc{next} for $T$ steps. We denote the sequence of states, actions, and rewards generated in such a way by $\{s_h\}_{h = 0}^{T}$, $\{a_h\}_{h = 0}^{T}$, and $\{r_h\}^{T}_{h = 0}$ respectively, where $s_h$ is the current state after calling \textsc{next} $h$ times, $a_h$ is the action taken at state $s_h$, and $r_h$ is the reward received after taking action $a_h$. 

For any $\gamma\in[0, 1)$, state $s$, and policy $\pi: \mathcal{S}\to\Delta(\mathcal{A})$, denote by $V^{\pi, \gamma}_s$ the expected  $\gamma$-discounted total rewards generated by starting from a state $s$ and following policy $\pi$ to choose the next action repeatedly, i.e. 
\begin{align*}
    V^{\pi, \gamma}_s = \expt_{\substack{s'_0 = s\\a'_h \sim \pi(s'_h)\\r'_h\sim R(s'_h, a'_h)\\ s'_{h + 1}\sim M(s'_h, a'_h)}}\left[\sum_{h = 0}^{\infty}\gamma^{h}r'_{h}\right].
\end{align*}
We can define the maximum $\gamma$-discounted total rewards from state $s$ by
\begin{align*}
    V^{*, \gamma}_{s} = \sup_{\pi}V^{\pi, \gamma}_{s}.
\end{align*}
The $\gamma$-regret is defined by
\begin{align*}
\Regret_\gamma(T) &= \sum_{h = 0}^{T - 1}(1 - \gamma)V^{*, \gamma}_{s_h} - \sum_{h = 0}^{T - 1}r_h.
\end{align*}

\section{Related Work}
Theoretical analysis on non-episodic RL are typically done through the notion of \emph{average-reward regret}, which is defined by
\begin{align*}
\Regret_{*}(T) &= \sum_{h = 0}^{T - 1}{\rho}^{*}_{s_h} - \sum_{h = 0}^{T - 1}r_h,
\end{align*}
where $\rho^*_{s_h}$ is the maximal \emph{gain} that can be achieved by a (stationary) policy starting from state $s_h$. For a more detailed introduction, see e.g.~\citet{mahadevan1996average}. It can be shown that the average-reward regret can be related to $\gamma$-regret by
\begin{align*}
    \lim_{\gamma\to 1}\Regret_{\gamma}(T) = \Regret_*(T)
\end{align*}
Current analysis of $\Regret_*(T)$ all assume that the MDP is at least weakly communicating, and therefore $\rho^*_{s_h}$ does not depend on $s_h$. The analysis was pioneered by~\citet{jaksch2010near}, who identified the diameter of the MDP, $D$, or any related measure, to be necessary when bounding $\Regret_*(T)$, and provided a lower bound of $\Omega(\sqrt{DSAT})$, which is still the best lower bound to date. They also derived an upper bound of $\tilde{O}(DS\sqrt{AT})$.

The definition of $\gamma$-regret is also quite related to a notion called \emph{sample complexity of exploration}, introduced in~\citet{kakade2003sample}. Specifically, this complexity, which we denote by $N_{\gamma}(\epsilon, \delta)$, is the smallest integer such that with probability at least $1 - \delta$, there are at most $N_{\gamma}(\epsilon, \delta)$ different $h$ such that $\mathbb{E}[\Delta_h] > \epsilon$, where
\begin{align*}
    \Delta_h = V^{*, \gamma}_{s_h} - \sum_{t = 0}^{\infty}\gamma^t r_{h + t}.
\end{align*}
The quantity $\Delta_h$ can be related to $\gamma$-regret by noting that
\begin{align*}
   (1 - \gamma)\sum_{h = 0}^{T - 1}\Delta_h - \frac{1}{1 - \gamma}\leq \Regret_{\gamma}(T) \leq (1 - \gamma)\sum_{h = 0}^{T - 1}\Delta_h + \frac{1}{1 - \gamma}. 
\end{align*}
However, $N_{\gamma}(\epsilon, \delta)$ itself does not measure directly the performance of a learner in the first $T$ steps. For example, a $T$-step optimal learner could have very large $N_{\gamma}(\epsilon, \delta)$ simply because it is not optimal after $T$ steps. The best upper bound on $N_{\gamma}(\epsilon, \delta)$ to date is $\tilde{O}\left(\frac{SA}{\epsilon^2(1 - \gamma)^6}\right)$~\cite{Szita+Szepesvari:2010}, while the best lower bound is $\tilde{\Omega}\left(\frac{SA}{\epsilon^2(1 - \gamma)^3}\right)$~\cite{lattimore2012pac}. More discussion on the connection between $N_{\gamma}(\epsilon, \delta)$ and $\Regret_{\gamma}(T)$ can be found in Section~\ref{sec: conn}.

The algorithm we use to prove the upper bounds on $\gamma$-regret is an adaption of~\citet[Algorithm~1]{jin2018q}, with modifications to handle the cyclic dependency in the non-episodic setting. For a detailed comparison see Sectiom~\ref{sec: rel-2}. It also looks visually similar to~\citet[Algorithm~1]{dong2019q}; however, the goal of~\citet{dong2019q} is to propose a model-free algorithm that has low sample complexity of exploration, and the proof technique therein is very different from ours and~\citet{jin2018q}.

\section{Lower Bounds}
In this section, we give two lower bounds on $\gamma$-regret. They both have $\sqrt{T}$ dependency on $T$ for sufficiently large $T$. The first one scales with $\sqrt{S}$ but has has a worse dependency on $\gamma$; the second one does not scale with $S$ but has a better dependency on $\gamma$. The proofs can be found in the appendix.
\begin{theorem}
\label{thm: lb1}
For any positive integers $S, A, T$, $\gamma\in[0, 1)$, and any (possibly randomized) learning algorithm, there exists an MDP such that 
\begin{align*}
    \expt\left[\Regret_{\gamma}(T)\right] \geq 
    \begin{dcases}
       \frac{T}{20}, &\text{if $T \leq SA$,}\\
       \frac{\sqrt{SAT}}{20\sqrt{2}},&\text{otherwise.}
       \end{dcases}
\end{align*}
\end{theorem}
\begin{theorem}
\label{thm: lb2}
For any $\gamma\in\left(\frac{2}{3}, 1\right)$, positive integers $A \geq 30, T \geq \frac{A}{1 - \gamma}$, and any (possibly randomized) learning algorithm, there exists a two-state MDP such that
\begin{align*}
    \mathbb{E}[\Regret_{\gamma}(T)] \ge\frac{\sqrt{AT}}{2304(1 - \gamma)^{\frac{1}{2}}}-\frac{1}{1 - \gamma}.
\end{align*}
\end{theorem}
\section{Upper Bounds}
\label{sec: sto-upper}
In this section, we will introduce a tabular version of the double Q-learning algorithm proposed in \citet{hasselt2010double}, and then use it to prove an upper bound on the $\gamma$-Regret. 

\begin{algorithm}[tb]
   \caption{Tabular Double Q-Learning with Upper Confidence}
   \label{alg: ucb-hoeffding}
\begin{algorithmic}
    \STATE {\bfseries Input:} $\gamma$, $\mathcal{S}$, $\mathcal{A}$, $p$ 
    \STATE {\bfseries Parameters:}  
    $b_t = \frac{2}{1 - \gamma} \sqrt{\frac{\ln\left(\frac{\pi^2SAt^2}{p}\right)}{(1 - \gamma)t}}$ for $t\geq 1$, $\alpha_t = \frac{2 - \gamma}{1 + t - t\gamma}$ for $t\geq 1$ 
    \vspace{0.8em}
    \FOR[start initialization]{$(s, a)\in\mathcal{S}\times\mathcal{A}$} 
        \STATE $Q^0(s, a), Q^1(s, a) \gets \frac{1}{1 - \gamma}$
        \STATE $N^0(s, a), N^1(s, a) \gets 0$
    \ENDFOR
    \STATE $s_0\gets\textsc{reset}$
    \FOR[main loop]{$h \gets 0, 1, \cdots, \infty$} 
        \STATE $\iota\gets h\bmod 2$
        \STATE $a_h\gets\argmax_{a'\in\mathcal{A}}Q^{\iota}(s_h, a')$ 
        \STATE $s_{h + 1}, r_h\gets \textsc{next}(a_h)$
        \STATE $N^{\iota}(s_h, a_h)\gets N^{\iota}(s_h, a_h) + 1$
        \STATE $t\gets N^{\iota}(s_h, a_h)$
       \State \begin{varwidth}[t]{\linewidth}
          $Q^{\iota}(s_h, a_h)\gets (1 - \alpha_t)Q^{\iota}(s_h, a_h)$\par
            \hskip\algorithmicindent 
            \hspace{4em}$ + \alpha_t \left(r + b_t +  \gamma\max_{a'\in\mathcal{A}}Q^{1 - \iota}(s_{h + 1}, a')\right)$
        \end{varwidth} 
    \ENDFOR
\end{algorithmic}
\end{algorithm}

The tabular version is described in Algorithm~\ref{alg: ucb-hoeffding}. Compared to
the original neural-network-oriented version, the tabular version has certain specifications that are crucial for theoretical analysis: 
\begin{itemize}
\item
The two Q-value functions have to be initialized to $\frac{1}{1 - \gamma}$, or for that matter, the maximal possible discounted cumulative return if it is known. 
\item 
The behavioral policy (i.e., the strategy used to choose which action to take) cannot be arbitrary as in the original version. The action has to be taken greedily based on the Q-value function that immediately gets updated afterward. 
\item In the original version, in each round, one of the two $Q$-value functions gets chosen, for example, randomly, and gets updated from the other one. In the tabular version, the updates have to happen in a strictly alternating fashion --- the $Q$-value function gets updated in a round is used for updating the other $Q$-value function in the next round. 
\item Soft update and upper confidence bounds are used when updating the $Q$-value functions.
\end{itemize}
Our upper bounds are presented in the following theorem.

\begin{theorem}
\label{thm: ucb-hoeffding}
For any $\gamma\in[0, 1)$ and $p\in(0, 1]$, with probability at least $1 - p$,  for any positive integer $T$, Algorithm~\ref{alg: ucb-hoeffding} has \[\Regret_{\gamma}(T) \leq \frac{14\sqrt{SAT\ln\left(\frac{\pi^2 SA T^2}{p}\right)}}{(1 - \gamma)^{\frac{3}{2}}} + \frac{2SA + 4}{1 - \gamma},\] and consequently,  
\begin{align*}
    \expt\left[\Regret_{\gamma}(T)\right] \leq \frac{14\sqrt{SAT\ln\left(\pi^2 SA T^3\right)}}{(1 - \gamma)^{\frac{3}{2}}} + \frac{2SA + 5}{1 - \gamma}.
\end{align*}
\end{theorem}
\subsection{Proof of Theorem~\ref{thm: ucb-hoeffding}}
The proof will be in the same style as in \citet{jin2018q}, with technical modifications to handle the cyclic dependencies in the non-episodic setting. Recall that in Algorithm~\ref{alg: ucb-hoeffding}, for any $t \geq 1$, $\alpha_t = \frac{2 - \gamma}{1 + t - t\gamma}$. We furthermore define $\alpha_0 = 1$. Let $\alpha_t^i = \alpha_i\prod_{j = i + 1}^t(1 - \alpha_j)$; it is easy to verify that $\sum_{i = 0}^t \alpha_t^i = 1$. Define by $Q_h$ and $N_h$ the $Q^0$ and $N^0$ function at the beginning of iteration $h$ if $h$ is even, or the $Q^1$ and the $N^1$ function at the beginning of iteration $h$ if $h$ is odd. Let $n_h = N_h(s_h, a_h)$. For $i = 1, 2, \cdots, n_h$, let $\prev_i(h)$ be the $i_{\text{th}}$ smallest $h' < h$ such that $h'$ and $h$ have the same parity, $s_{h'} = s_{h}$, and $a_{h'} = a_{h}$. Define 
\begin{align*}
    &V_h(s) = \max_a Q_h(s, a), &&\Delta_h = V^{*, \gamma}_{s_h} - \sum_{t = 0}^{\infty}\gamma^t r_{h + t},\\
    &\bar{R}(s, a) = \expt_{r\sim R(s, a)}[r], 
    &&\bar{r}_h = \bar{R}(s_h, a_h),\\
    &V^{*, \gamma}_{M(s, a)} = \expt_{s'\sim M(s, a)}\left[V^{*, \gamma}_{s'}\right], 
    &&Q^{*, \gamma}_{s, a} = \bar{R}(s, a) + V^{*, \gamma}_{M(s, a)},\\
    &\phi_h = V_h(s_h) - V^{*, \gamma}_{s_h}, &&\delta_h = \phi_h + \Delta_h,
\end{align*}
The following lemmas are adapted from \citet{jin2018q}; the proofs can be found in the appendix. 
\begin{lemma}
The following statements are true:
\label{lem: basics}
       (i). 
        $\frac{\ln(C\cdot t)}{\sqrt{t}}
        \leq \sum_{i = 1}^t\alpha^i_t\sqrt{\frac{\ln(C\cdot i)}{i}}\leq 2\cdot\frac{\ln(C\cdot t)}{\sqrt{t}}$\,\,   for any $t\geq 1$ and $C\geq e$;
        (ii).
        $\sum_{i = 1}^{t}\left(\alpha^i_t\right)^2 \leq \frac{2}{(1 - \gamma)t}$\,\,  for any $t\geq 1$;
        (iii). $\sum_{t = i}^{\infty}\alpha^i_t = 2 - \gamma$\,\,  for any $i\geq 1$.
\end{lemma}
\begin{lemma}
\label{lem: diff-expansion}
    For any $h$, 
    \begin{align*}
        Q_h(s_h, a_h) - Q^{*, \gamma}_{s_h, a_h} 
    &=\alpha_{n_h}^0\left(\frac{1}{1 - \gamma} - Q_{s, a}^{*, \gamma}\right) 
    + \sum_{i = 1}^{n_h}\alpha_{n_h}^i b_i + \gamma\sum_{i = 1}^{n_h}\alpha_{n_h}^i \phi_{\prev_h^i + 1} \\
    &\ \ \ \ +\sum_{i = 1}^{n_h}\alpha_{n_h}^i\left(r_{\prev^i_h} - \bar{r}_{\prev_h^i} + \gamma\left(V^{*, \gamma}_{s_{\prev_h^i + 1}} - V^{*, \gamma}_{M(s_h, a_h)}\right)\right).
    \end{align*}
\end{lemma}
\begin{lemma}
\label{lem: diff-bound}
Define random variables $r_{s, a, i}$ to be the reward received after taking action $a$ on state $s$ the $i_{\text{th}}$ time, and $s'_{s, a, i}$ to be the next state when receiving reward $r_{s, a, i}$, then for any $T$, with probability at least $1 - p$, the following hold simultaneously
\begin{itemize}
\item[(i).]
For any $h$, 
    $0 \leq Q_h(s_h, a_h) - Q^{*, \gamma}_{s_h, a_h} \leq \frac{\alpha_{n_h}^0}{1 - \gamma}  
    + \gamma\sum_{i = 1}^{n_h}\alpha_{n_h}^i \phi_{\prev_h^i + 1} + 3\beta_{n_h}$, 
where $\beta_{t} = \frac{2}{1 - \gamma}\sqrt{\frac{\ln\left(\frac{\pi^2SA t^2}{p}\right)}{(1 - \gamma)t}}$ if $t\geq 1$ and $\beta_0 = 0$.
\item[(ii).]
$
\sum_{h = 0}^{T - 1}\bar{r}_{h} - r_h  + \gamma\left(V^{*, \gamma}_{M(s_h, a_h)} - V^{*, \gamma}_{s_{h + 1}}\right) \leq \xi_T$,
where $\xi_T = \frac{\sqrt{2}}{1 - \gamma} \sqrt{T\ln\left(\frac{3}{2p}\right)}$.
\end{itemize}
\end{lemma}
We are now ready to begin our proof. From now on all the calculation will condition on the events where the statements in Lemma~\ref{lem: diff-bound} are true. It is important to notice that in this case we have that for any $h$, $\phi_h \geq 0$ and $\Delta_h\leq \delta_h$. First note that
\begin{align*}
    \delta_h &= Q_h(s_h, a_h) -\sum_{t = 0}^{\infty}\gamma^t r_{h + t}\\
    &= \left(Q_h(s_h, a_h) - Q^{*, \gamma}_{s_h, a_h}\right) + \left(Q^{*, \gamma}_{s_h, a_h} - \sum_{t = 0}^{\infty}\gamma^t r_{h + t}\right)\\
    &\stackrel{(a)}{\leq}
    \alpha_{n_h}^0\cdot\frac{1}{1 - \gamma}  
    + \gamma\sum_{i = 1}^{n_h}\alpha_{n_h}^i \phi_{\prev_h^i + 1} +3\beta_{n_h} + \bar{r}_{h} - r_h + \gamma\left(V^{*, \gamma}_{M(s_h, a_h)} - \sum_{t = 0}^{\infty}\gamma^t r_{h + 1 + t}\right)\\
    &\leq \alpha_{n_h}^0\cdot\frac{1}{1 - \gamma}  
    + \gamma\sum_{i = 1}^{n_h}\alpha_{n_h}^i \phi_{\prev_h^i + 1} +3\beta_{n_h}+ 
    \gamma\left(\delta_{h + 1}- \phi_{h + 1}\right) + \left(\bar{r}_{h} - r_h\right)   + \gamma\left(V^{*, \gamma}_{M(s_h, a_h)} - V^{*, \gamma}_{s_{h + 1}}\right),
\end{align*}
where (a) is due to Lemma~\ref{lem: diff-bound}.(i). Therefore, according to Lemma~\ref{lem: diff-bound}.(ii) we have
\begin{align*}
\sum_{h = 0}^{T - 1}\delta_h \leq \frac{1}{1 - \gamma}\sum_{h = 0}^{T - 1}\alpha_{n_h}^0 + \gamma\sum_{h = 0}^{T - 1}\sum_{i = 1}^{n_h}\alpha_{n_h}^i\phi_{\prev_h^i + 1} + \gamma\sum_{h = 0}^{T - 1}\delta_{h + 1} - \gamma\sum_{h = 0}^{T - 1}\phi_{h + 1} + \xi_T + 3\sum_{h = 0}^{T - 1}\beta_{n_h}.
\end{align*}
Using the fact that $\lvert\delta_h\rvert \leq \frac{1}{1 - \gamma}$ for any $h$ and rearranging the terms, we get
\begin{align}
\label{eq: goback}
(1 - \gamma)\sum_{h = 0}^{T - 1}\delta_h &\leq \frac{1}{1 - \gamma}\sum_{h = 0}^{T - 1}\alpha_{n_h}^0 + \gamma\left(\sum_{h = 0}^{T - 1}\sum_{i = 1}^{n_h}\alpha_{n_h}^i\phi_{\prev_h^i + 1} - \sum_{h = 0}^{T - 1}\phi_{h + 1}\right) + \xi_T + 3\sum_{h = 0}^{T - 1}\beta_{n_h} + \frac{2\gamma}{1 - \gamma}.
\end{align}
To continue the calculation, first note that $\alpha^0_{n_h}$ is $1$ if $n_h = 0$ and is $0$ otherwise, therefore \[\sum_{h = 0}^{T - 1}\alpha^0_{n_h} = \sum_{h = 0}^{T - 1}\mathbbm{1}_{n_h = 0} \leq 2SA.\] Next note that
\begin{align*}
     \sum_{h = 0}^{T - 1}\sum_{i = 1}^{n_h}\alpha_{n_h}^i\phi_{\prev_h^i + 1} - \sum_{h = 0}^{T - 1}\phi_{h + 1}&\leq \sum_{h = 0}^{T - 1}\phi_h\sum_{t = n_h + 1}^{\infty}\alpha^{n_h}_{t} - \sum_{h = 0}^{T - 1}\phi_{h + 1}\\
     &\stackrel{(a)}{\leq} \frac{2 - \gamma}{2}\sum_{h = 0}^{T - 1}\phi_h  - \sum_{h = 0}^{T - 1}\phi_{h} + \phi_0\\
     &=(1 - \gamma)\sum_{h = 0}^{T - 1}\phi_{h} + \frac{1}{1 - \gamma}\\
     &=(1 - \gamma)\sum_{h = 0}^{T - 1}\left(\delta_{h} - \Delta_h\right) + \frac{1}{ 1- \gamma},
\end{align*}
where (a) is because of Lemma~\ref{lem: basics}.(iii). Now going back to \eqref{eq: goback} we get
\begin{align*}
    &(1 - \gamma)\sum_{h = 0}^{T - 1}\delta_h \leq  \frac{2SA + 3}{1 - \gamma} +(1 - \gamma)\sum_{h = 0}^{T - 1} (\delta_h - \Delta_h)+  \xi_T + 3\sum_{h = 0}^{T - 1}\beta_{n_h}\\
    &\iff 
    (1 - \gamma)\sum_{h = 0}^{T - 1}\Delta_h \leq  \frac{2SA + 3}{1 - \gamma} +  \xi_T + 3\sum_{h = 0}^{T - 1}\beta_{n_h}\\
\end{align*}
Finally, note that 
\begin{align*}
    \sum_{h = 0}^{T - 1}\beta_{n_h} &\leq
    \frac{2\sqrt{\ln\left(\frac{\pi^2 SA T^2}{p}\right)}}{(1 - \gamma)^{1.5}}\sum_{h = 0}^{T - 1}\mathbbm{1}_{n_h \geq 1}\cdot\sqrt{\frac{1}{n_h}}\\
    &=\frac{2\sqrt{\ln\left(\frac{\pi^2 SA T^2}{p}\right)}}{(1 - \gamma)^{1.5}}\sum_{s, a}\sum_{t = 1}^{N_{T}(s, a)}\sqrt{\frac{1}{t}}\\
    &\stackrel{(a)}{\leq}\frac{2\sqrt{\ln\left(\frac{\pi^2 SA T^2}{p}\right)}}{(1 - \gamma)^{1.5}}\sum_{s, a}2\sqrt{N_T(s, a)}\\
    &\stackrel{(b)}{\leq}\frac{4\sqrt{SAT\ln\left(\frac{\pi^2 SA T^2}{p}\right)}}{(1 - \gamma)^{1.5}},
\end{align*}
where (a) is because $\sum_{i = 1}^t \sqrt{\frac{1}{i}} \leq 2\sqrt{t}$ and (b) is by Cauchy-Schwarz inequality and the fact that $\sum_{s, a}N_T(s, a)\leq T$. Therefore,
\begin{align}
    (1 - \gamma)\sum_{h = 0}^{T - 1}\Delta_h &\leq  \frac{2SA + 3}{1 - \gamma} +  \frac{\sqrt{2}}{1 - \gamma}\sqrt{T\ln\left(\frac{3}{2p}\right)} + \frac{12\sqrt{SAT\ln\left(\frac{\pi^2 SA T^2}{p}\right)}}{(1 - \gamma)^{1.5}}\nonumber\\
    &\leq  \frac{2SA + 3}{1 - \gamma} +   \frac{14\sqrt{SAT\ln\left(\frac{\pi^2 SA T^2}{p}\right)}}{(1 - \gamma)^{1.5}}\label{eq: comb1}
\end{align}
On the other hand, we have
\begin{align}
(1 - \gamma) \sum_{h = 0}^{T - 1}\Delta_h &= \sum_{h = 0}^{T - 1}(1 - \gamma)V^{*, \gamma}_{s_h} - (1 - \gamma)\sum_{h = 0}^{T - 1}\sum_{t = 0}^{\infty}\gamma^t r_{h + t}\nonumber\\ 
&\geq \sum_{h = 0}^{T - 1}(1 - \gamma)V^{*, \gamma}_{s_h} - \sum_{u=0}^{T - 1}r_u - (1 - \gamma)\sum_{u = T}^{\infty}r_u\sum_{v = u - T + 1}^{u}\gamma^v\nonumber\\
&=\Regret_{\gamma}(T) - (1 - \gamma)\sum_{u = T}^{\infty}r_u\sum_{v = u - T + 1}^{u}\gamma^v\nonumber\\
&\geq \Regret_{\gamma}(T) - \frac{1}{1 - \gamma}.\label{eq: comb2}
\end{align}
Combining together \eqref{eq: comb1} and \eqref{eq: comb2}, we arrive at
\begin{align*}
    \Regret_{\gamma}(T) \leq   \frac{14\sqrt{SAT\ln\left(\frac{\pi^2 SA T^2}{p}\right)}}{(1 - \gamma)^{1.5}} + \frac{2SA + 4}{1 - \gamma}.
\end{align*}
This concludes the proof.

\section{Technical Discussions}

\subsection{Converting Sample Complexity of Exploration to $\gamma$-Regret}
\label{sec: conn}
Recall that the sample complexity of exploration $N_{\gamma}(\epsilon, \delta)$ is the smallest integer such that with probability at least $1 - \delta$, there are at most $N(\epsilon, \delta)$ different $h$ such that
\begin{align*}
    V^{*, \gamma}_{s_h} - \expt\left[\sum_{t = 0}^{\infty}\gamma^t r_{h + t}\right] > \epsilon.
\end{align*}

It is easy to see that
\begin{align*}
   \expt\left[\Regret_{\gamma}(T)\right]\in O\left(\inf_{\epsilon}N_{\gamma}\left(\epsilon, \frac{1}{T}\right) + \epsilon T (1 - \gamma) + \frac{1}{1 - \gamma}\right).
\end{align*}
Plugging in the best existing bound for $N_{\gamma}(\epsilon, \delta)$, which is $\tilde{O}\left(\frac{SA}{\epsilon^2(1 - \gamma)^6}\right)$ from~\citet{Szita+Szepesvari:2010}, we arrive at an upper bound of $\tilde{O}\left(\frac{T^{\frac{2}{3}}(SA)^{\frac{1}{3}}}{(1 - \gamma)^{\frac{4}{3}}}\right)$ on $\expt\left[\Regret_\gamma(T)\right]$. It may seem that this bound has better dependencies on $S$, $A$, and $\gamma$, but this is not the case. In fact, we have the following inequalities:
\begin{align}
    \tilde{O}\left(\frac{T^{\frac{2}{3}}(SA)^{\frac{1}{3}}}{(1 - \gamma)^{\frac{4}{3}}}\right) \geq 
    \begin{dcases}
    \tilde{O}\left(T\right), &\hspace{-0.3em}\text{if $T < \frac{SA}{(1 - \gamma)^4}$,}\\
    \tilde{O}\left(\frac{\sqrt{SAT}}{(1 - \gamma)^{2}}\right), &\hspace{-0.3em}\text{otherwise.}
    \end{dcases}
    \label{eq: transform}
\end{align}
Note that in the above inequalities $\tilde{O}\left(T\right)$ is a trivial upper bound on $\Regret_{\gamma}(T)$ for any $T$, while $\tilde{O}\left(\frac{\sqrt{SAT}}{(1 - \gamma)^{2}}\right)$ has a worse dependency on $\gamma$ than our upper bound. 

If the upper bounds on $N_{\gamma}(\epsilon, \delta)$ were to hold uniformly over all possible $\epsilon$, then we could translate the (uniform) upper bound on $N_{\gamma}(\epsilon, \delta)$ into upper bounds on $\gamma$-regret in a better way. In fact, if the best existing upper bound on $N_{\gamma}(\epsilon, \delta)$, $\tilde{O}\left(\frac{SA}{\epsilon^2(1 - \gamma)^6}\right)$, was to hold uniformly over all possible $\epsilon$, then  
\begin{align*}
   \expt\left[\Regret_{\gamma}(T)\right]\in O\left((1 - \gamma)\left(\int_{\epsilon_0}^{\frac{1}{1 - \gamma}}\frac{SA}{\epsilon^2(1 - \gamma)^6} + T\epsilon_0\right) + \frac{1}{1 - \gamma}\right),
\end{align*}
we could get an upper bound on $\gamma$-regret as good as $\tilde{O}\left(\frac{\sqrt{SAT}}{(1 - \gamma)^{2}}\right)$. We can see that even in this imagined ideal scenario the translated upper bound still has a worse dependency on $\gamma$ than ours.

\subsection{Lower Bound Proof Techniques}
Our proof of the second lower bound (Theorem~\ref{thm: lb2}) on $\gamma$-regret is an adaptation of the proof for the average-reward setting in~\citet[Theorem~5]{jaksch2010near}. The major challenge the $\gamma$-regret formulation brings is that the value function, now being the discounted total return instead of the long-term average, can vary from state to state. While in the two-state MDP case this is still manageable by explicitly writing out the exact formula of the value functions for each state, as we have done in the proof of Theorem~\ref{thm: lb2}, it becomes unclear how we should generalize to MDPs with $S$ states. 

\subsection{Upper Bounds Proof Techniques}
\label{sec: rel-2}
Our derivation of the upper bounds on $\gamma$-regret is inspired by \citet{jin2018q}, who showed that a specific tabular version of Q-Learning~\cite{watkins1992q} has near-optimal regret in the episodic setting. Their analysis, however, is not directly applicable to the non-episodic setting:

First, in the episodic setting, there are $H$ value functions $Q_0, Q_1, \cdots, Q_{H - 1}$ to be learned, each $Q_i$ depends only on $Q_{j > i}$  --- there is no cyclic dependency; on the other hand, in the non-episodic setting, there is only one single value function, so a hierarchical induction in the analysis is not possible. To deal with self-dependency, we find it very useful to replace the regular Q-learning with double Q-learning~\cite{hasselt2010double}, which has been widely used in deep reinforcement learning since it was introduced~\cite{van2016deep, hessel2018rainbow}. 

Second, a key ingredient in the proof of \citet{jin2018q} is the choice of learning rate $\alpha_t = \frac{H + 1}{H + t}$ --- a nice consequence of this choice is that the total per-episode-step regret blow-up is $\left(1 + 1/H\right)$; since there are at most $H$ steps in each episode, the total blow-up is $\left(1 + 1/H\right)^H$, which is upper bounded by the constant $e$ regardless how large $H$ is. The same quantity $\left(1 + 1/H\right)^H$ also appeared in~\citet{azar2017minimax} for the same reason. However, in the non-episodic setting, the blow-up could become arbitrarily large because the learner is not reset every $H$ steps; therefore, different techniques are required to control the blow-up of the regret.

\bibliography{ref}
\newpage
\appendix

\section{Proof of Theorem~\ref{thm: lb1}}
    The proof will essentially be a reduction to the multi-armed bandit lower bounds. Fix $S$, $A$, $T$, $\gamma$, and a learning algorithm, we construct an MDP as follows: if we label all the states by $0, 1, \cdots, S - 1$, then the learner goes from state $i$ to state $((i + 1) \mod S)$ deterministically regardless the action taken, and the reward function is specified by 
    \begin{align*}
        R(s, a) = \begin{dcases}
        1, &\text{\ w.p. $\frac{1}{2} + \mathbbm{1}_{a = a_s}\cdot \epsilon$,}\\
        0, &\text{\ otherwise.}
        \end{dcases}
    \end{align*}
    where $\epsilon$ and $a_s$ will be specified later. It is easy to see that 
    \begin{align}
        \expt\left[\Regret_{\gamma}(T)\right] = \left(T\cdot\left(\frac{1}{2} + \epsilon\right) - \expt\left[\sum_{u = 0}^{T - 1}r_u\right] \right).
        \label{eq: retrace-2}
    \end{align}
    Let $T_s$ be the number of actions the learner takes at state $s$ in the first $T$ steps, we have that 
    \begin{align}
    T_s \geq 
     \left\lfloor\frac{T}{S}\right\rfloor\geq \frac{T}{2S}\ \ \text{if $T\geq S$}.\label{eq: Ts} 
    \end{align} 
    Also denote $r_{s, i}$ the $i_{\text{th}}$ reward received when taking action at state $s$, then we have that
    \begin{align}
        \expt\left[\Regret_{\gamma}(T)\right] &= \sum_{s = 1}^{S} 
        \left(T_s\cdot\left(\frac{1}{2}+\epsilon\right) - \expt\left[\sum_{v = 1}^{T_s}r_{s, v}\right]\right).\label{eq: retrace-1}
    \end{align}
    According to the lower bounds for multi-armed bandits, e.g. Theorem 7.1 and its construction in \citet{auer1995gambling}, there exists $\epsilon = \frac{1}{4}\min\left(\sqrt{\frac{A}{T}}, 1\right)$ and $a_s$ (recall that $R(s, a)$ is defined from $\epsilon$ and $a_s$) such that for any $s$ we have
    \begin{align*}
        T_s\cdot\left(\frac{1}{2}+\epsilon\right) - \expt\left[\sum_{v = 1}^{T_i}r_{s, v}\right] \geq \frac{1}{20}\min\left(\sqrt{AT_s}, T_s\right).
    \end{align*}
    Finally, note that if $T\leq SA$, then $T_s \leq A$ for any $s$ and consequently $\min\left(\sqrt{A T_s}, T_s\right) = T_s$; on the other hand, if $T > SA$, then $T_s \geq A$ for any $s$ and consequently for any $s$, $\min\left(\sqrt{A T_s}, T_s\right) = \sqrt{A T_s} \geq \sqrt{\frac{AT}{2S}}$, where the last inequality follows from \eqref{eq: Ts}. Going back to \eqref{eq: retrace-1} gives us the desired lower bounds.
    
\section{Proof of Theorem~\ref{thm: lb2}}
We will construct an MDP similar to the one in the proof of ~\citet[Theorem~5]{jaksch2010near} for our proof. Specifically, the MDP has two states $0$ and $1$; the learner receives reward $0$ in state $0$ and reward $1$ in state $1$, regardless the action taken; the learner goes from state $1$ to state $0$ with probability $1 - \gamma$ regardless the action taken; the learner goes from state $0$ to state $1$ with probability $1 - \gamma + \mathbbm{1}_{a=a^*}\cdot\epsilon$ when action $a$ is taken, where $\epsilon = \frac{1}{24}\sqrt{\frac{A(1 -\gamma)}{T}}$ and $a^*$ will be chosen later. It is easy to see that $\epsilon \leq 1 - \gamma$ since we assumed that $T \geq \frac{A}{1 - \gamma}$. By definition, we have that
    \begin{align*}
        V^{*, \gamma}_0 &= \gamma(1 - \gamma + \epsilon)V^{*, \gamma}_1 + \gamma(\gamma - \epsilon)V^{*, \gamma}_0,\\
        V^{*, \gamma}_1 &= 1 +  \gamma(1 - \gamma) V^{*, \gamma}_0 + \gamma^2 V^{*, \gamma}_1.
    \end{align*}
We can solve the above equations to get
    \begin{align}
        V^{*, \gamma}_0 &= \frac{\gamma - \gamma^2 + \gamma\epsilon}{(1 - \gamma)(1 - 2\gamma^2 + \gamma + \gamma \epsilon)},\label{d1}\\
        V^{*, \gamma}_1 &=  \frac{1 - \gamma^2 + \gamma\epsilon}{(1 - \gamma)(1 - 2\gamma^2 + \gamma + \gamma \epsilon)}.\label{d2}
    \end{align}
    Note that because $\epsilon \leq 1 - \gamma$, we have in the denominators of \eqref{d1} and \eqref{d2} that
    \begin{align}
         (1 - \gamma)(1 - 2\gamma^2 + \gamma + \gamma \epsilon)\in \left[(1 - \gamma)^2, 4(1 - \gamma)^2\right]\label{eq: twosidedeno}
    \end{align} 
    Let $N_0$ and $N_1$ be the number of steps (in the first $T$ steps) that the leaner is in state $0$ and $1$ respectively, and let $N_0^*$ be the number of steps (in the first $T$ steps) the learner is in state $0$ and takes action $a^*$, using the same argument as in the proof of~\citet[Theorem~5]{jaksch2010near}, we have that
    \begin{align}
        E[N_1] &\leq \frac{T}{2} + \mathbb{E}[N_0^*]\cdot\frac{\epsilon}{1 - \gamma} + \frac{1}{2(1 - \gamma)},\label{eq: N1}\\
        \mathbb{E}[N_0^*] &\leq \frac{T}{2A} + \frac{1}{2A(1 - \gamma)} + \frac{\epsilon T}{2}\sqrt{\frac{T}{A(1 - \gamma)}} + \frac{\epsilon T}{2(1 - \gamma)\sqrt{A}}.\label{eq: N0*}
    \end{align}
    Therefore, 
    \begin{align*}
        &(1 - \gamma)^{-1}\cdot\mathbb{E}[\Regret(T)]\\ &\geq \mathbb{E}[N_0]\cdot V_0^{*, \gamma} + \mathbb{E}[N_1] \cdot\left(V_1^{*, \gamma} - \frac{1}{1 - \gamma}\right)\\
        &=\frac{\mathbb{E}[N_0]\cdot\gamma(1 - \gamma + \epsilon) - \mathbb{E}[N_1]\cdot\gamma(1 - \gamma)}{(1 - \gamma)(1 - 2\gamma^2 + \gamma + \gamma \epsilon)} \\
        &\stackrel{(a)}{\geq}\frac{\frac{T\gamma\epsilon}{2} - \gamma - \frac{\epsilon\gamma}{2(1 - \gamma)} - \mathbb{E}[N_0^*]\cdot\frac{\epsilon\gamma(2 - 2\gamma + \epsilon)}{1 - \gamma}}{(1 - \gamma)(1 - 2\gamma^2 + \gamma + \gamma \epsilon)} \\
        &\stackrel{(b)}{\geq}\gamma\cdot\frac{\frac{T\epsilon}{2} - 1 - \frac{\epsilon}{2(1 - \gamma)} - \left(\frac{T}{2A} + \frac{1}{2A(1 - \gamma)} + \frac{\epsilon T}{2}\sqrt{\frac{T}{A(1 - \gamma)}} + \frac{\epsilon T}{2(1 - \gamma)\sqrt{A}}\right)\cdot\frac{\epsilon(2 - 2\gamma + \epsilon)}{1 - \gamma}}{(1 - \gamma)(1 - 2\gamma^2 + \gamma + \gamma \epsilon)} \\
        &\stackrel{(c)}{\geq}\gamma\cdot\frac{\frac{T\epsilon}{4} - 1 -  3\epsilon\cdot\left(\frac{5T}{8A} + \frac{\epsilon T}{2}\sqrt{\frac{T}{A(1 - \gamma)}} + \frac{\epsilon T}{2(1 - \gamma)\sqrt{A}}\right)}{(1 - \gamma)(1 - 2\gamma^2 + \gamma + \gamma \epsilon)}\\
        &\stackrel{(d)}{=}\gamma\cdot\frac{\frac{\sqrt{AT(1 - \gamma)}}{96} - 1 -  \left(\frac{\sqrt{AT(1 - \gamma)}}{384} + \frac{\sqrt{AT(1 - \gamma)}}{192} \right)}{(1 - \gamma)(1 - 2\gamma^2 + \gamma + \gamma \epsilon)} \\
        &\stackrel{(e)}{=}\frac{\sqrt{AT(1 - \gamma)}}{576(1 - \gamma)(1 - 2\gamma^2 + \gamma + \gamma \epsilon)} - \frac{1}{(1 - \gamma)(1 - 2\gamma^2 + \gamma + \gamma \epsilon)} \\
        &\stackrel{(f)}{\geq}\frac{\sqrt{AT}}{2304(1 - \gamma)^{1.5}}-\frac{1}{(1 - \gamma)^2},
    \end{align*}
    where (a) is due to \eqref{eq: N1} and the fact that $N_0 + N_1 = T$, (b) is due to \eqref{eq: N0*}, (c) is because by assumption $T\geq \frac{A}{1 - \gamma} \geq \frac{4}{1 - \gamma}$ and $\epsilon \leq 1 - \gamma$, (d) is by substituting $\epsilon$ with the chosen value and recall that by assumption $A\geq 30$ and $T(1 - \gamma) \geq 1$, (e) is because by our assumption $\gamma\in \left(\frac{2}{3}, 1\right)$, (f) is due to \eqref{eq: twosidedeno}. Rearranging the terms concludes the proof.

\section{Proof of Lemma~\ref{lem: basics}}
For (ii) and (iii), the same proof as in \citet{jin2018q}, Lemma 4.1.(b)-(c) can be applied, with $H$ replaced by $\frac{1}{1 - \gamma}$, and note that in proving (iii) the requirement for $n$ and $k$ to be positive integers in their proof can be relaxed to $n$ and $k$ being real numbers that are at least $1$. We will prove (i) by induction on $t$. The base case $t = 1$ holds because $\alpha_t^1 = 1$. Assuming the statement is true for $t$, then on one hand, 
\begin{align*}
    \sum_{i = 1}^{t + 1}\alpha_{t + 1}^i\sqrt{\frac{\ln\left(C\cdot i\right)}{i}} 
    &\stackrel{(a)}{=} \alpha_{t + 1}\sqrt{\frac{\ln\left(C\cdot (t + 1)\right)}{t + 1}} + (1 - \alpha_{t + 1})\sum_{i = 1}^{t }\alpha^i_{t}\sqrt{\frac{\ln\left(C\cdot i\right)}{i}}\\
    &\stackrel{(b)}{\geq} \alpha_{t + 1}\sqrt{\frac{\ln\left(C\cdot (t + 1)\right)}{t + 1}} + (1 - \alpha_{t + 1})\sqrt{\frac{\ln\left(C\cdot t\right)}{t}}\\
    &\stackrel{(c)}{\geq}\sqrt{\frac{\ln\left(C\cdot (t + 1)\right)}{t + 1}},
\end{align*}
where in (a) we used the definition of $\alpha^i_t$, in (b) we used the induction assumption, and (c) is because $x\mapsto \frac{\ln(C\cdot x)}{x}$ is a non-increasing function when $C\geq e$ and $x \geq 1$. On the other hand, we have
\begin{align*}
    \sum_{i = 1}^{t + 1}\alpha_{t + 1}^i\sqrt{\frac{\ln\left(C\cdot i\right)}{i}} 
    &\stackrel{(a)}{=} \alpha_{t + 1}\sqrt{\frac{\ln\left(C\cdot (t + 1)\right)}{t + 1}} + (1 - \alpha_{t + 1})\sum_{i = 1}^{t }\alpha^i_{t}\sqrt{\frac{\ln\left(C\cdot i\right)}{i}}\\
    &\stackrel{(b)}{\leq} \alpha_{t + 1}\sqrt{\frac{\ln\left(C\cdot (t + 1)\right)}{t + 1}} + 2(1 - \alpha_{t + 1})\sqrt{\frac{\ln\left(C\cdot t\right)}{t}}\\
    &\stackrel{(c)}{=} \frac{2 - \gamma}{2 + t - (t + 1)\gamma}\sqrt{\frac{\ln\left(C\cdot (t + 1)\right)}{t + 1}} + \frac{2t(1 - \gamma)}{2 + t - (t + 1)\gamma}\sqrt{\frac{\ln\left(C\cdot t\right)}{t}}\\
    &\leq \frac{2 - \gamma}{2 + t - (t + 1)\gamma}\sqrt{\frac{\ln\left(C\cdot (t + 1)\right)}{t + 1}} + \frac{2\sqrt{t}(1 - \gamma)\sqrt{t + 1}}{2 + t - (t + 1)\gamma}\sqrt{\frac{\ln\left(C\cdot (t + 1)\right)}{t + 1}}\\
    &\leq \frac{2 + 2(t + 1)(1 - \gamma)}{1 + (t + 1)(1 - \gamma)}\sqrt{\frac{\ln\left(C\cdot (t + 1)\right)}{t + 1}}\\
    &= 2\sqrt{\frac{\ln\left(C\cdot (t + 1)\right)}{t + 1}}.
\end{align*}
where in (a) we used the definition of $\alpha^i_t$, in (b) we used the induction assumption, and in (c) we used the definition of $\alpha_t$. Therefore, the statement in (i) is true for any $t\geq 1$.
First note that
\section{Proof of Lemma~\ref{lem: diff-expansion}}
We have that
\begin{align*}
    Q_h(s_h, a_h) &= \alpha_{n_h}^0\frac{1}{1 - \gamma} + \sum_{i = 1}^{n_h}\alpha^i_{n_h}\left(r_{\prev_h^i} + b_i + \gamma V_{\prev_h^i + 1}\left(s_{\prev_h^i + 1}\right)\right)\\
    Q^{*, \gamma}_{s_h, a_h} &\stackrel{(a)}{=} \sum_{i = 0}^{n_h}\alpha_{n_h}^i \left(\bar{r}_{\prev^i_h} + \gamma V^{*, \gamma}_{M(s_h, a_h)}\right) = \alpha_{n_h}^0 Q^{*, \gamma}_{s, a} + \sum_{i = 1}^{n_h}\alpha_{n_h}^i\left(\bar{r}_{\prev^i_h} + \gamma V^{*, \gamma}_{M(s_h, a_h)}\right),
\end{align*}
where in (a) we used the fact that $\sum_{i = 0}^t\alpha_t^i = 1$ for any $t$ and the definition of $Q^{*, \gamma}_{s, a}$. Therefore we have
    \begin{align*}
        Q_h(s_h, a_h) - Q^{*, \gamma}_{s_h, a_h} &= \alpha_{n_h}^0\left(\frac{1}{1 - \gamma} - Q^{*, \gamma}_{s, a}\right)  
    + \sum_{i = 1}^{n_h}\alpha_{n_h}^i\left(r_{\prev^i_h} - \bar{r}_{\prev^i_h}+ b_i + \gamma\left(V_{\prev^i_h + 1}(s_{\prev^i_h + 1}) - V^{*, \gamma}_{M(s_h, a_h)}\right)\right)\\
    &=\alpha_{n_h}^0\left(\frac{1}{1 - \gamma} - Q_{s, a}^{*, \gamma}\right) 
    + \sum_{i = 1}^{n_h}\alpha_{n_h}^i b_i + \gamma\sum_{i = 1}^{n_h}\alpha_{n_h}^i \phi_{\prev_h^i + 1} \\
    &+\sum_{i = 1}^{n_h}\alpha_{n_h}^i\left(r_{\prev^i_h} - \bar{r}_{\prev_h^i} + \gamma\left(V^{*, \gamma}_{s_{\prev_h^i + 1}} - V^{*, \gamma}_{M(s_h, a_h)}\right)\right).
    \end{align*}
    
\section{Proof of Lemma~\ref{lem: diff-bound}}
It suffices to show that (i) happens with probability at least $1 - \frac{p}{3}$ and (ii) happens with probability at least $1 - \frac{2p}{3}$. 

We focus on (i) first.  
The case where $n_h = 0$ is trivial, so we assume $n_h \geq 1$. Fix any $s, a, t$, let $x_i = \alpha_{t}^i\left(r_{s, a, i} - \bar{R}(s, a) + \gamma\left(V^{*, \gamma}_{s'_{s, a, i}} - V^{*, \gamma}_{M(s, a)}\right)\right)$. We can see that $\left\{x_i\right\}_{i = 1}^{t}$ is a Martingale difference sequence and $\lvert x_i\rvert \leq  \frac{\alpha_t^i}{1 - \gamma}$, therefore by Azuma-Hoeffding inequality we have that with probability at least $1 - \frac{2p}{\pi^2 SAt^2}$,
\begin{align*}
    \left\lvert\sum_{i = 1}^t x_i\right\rvert
    &\leq\frac{1}{1 - \gamma} \sqrt{2\ln\left(\frac{\pi^2SAt^2}{p}\right)\sum_{i = 1}^t\left(\alpha_t^i\right)^2}\\
    &\stackrel{(a)}{\leq}\frac{1}{1 - \gamma} \sqrt{\frac{4\ln\left(\frac{\pi^2SAt^2}{p}\right)}{(1 - \gamma)t}} = \beta_t,
\end{align*}
where in (a) we used Lemma~\ref{lem: basics}.(ii).
Using a union bound, the above inequalities hold for all $s$, $a$, $t\geq 1$ with probability at least 
\begin{align*}
    1 - SA\sum_{t = 1}^{\infty}\frac{2p}{\pi^2SAt^2} = 1 - \frac{p}{3}.
\end{align*}
According to Lemma~\ref{lem: diff-expansion}, it suffices to show that with probability at least $1 - \frac{p}{3}$, we have that for any $s, a, t$, 
\begin{align}
0\leq\sum_{i = 1}^{t}\alpha_{t}^i b_i 
    +\sum_{i = 1}^{t}\alpha_{t}^i\left(r_{s, a, i} - \bar{R}(s, a) + \gamma\left(V^{*, \gamma}_{s'_{s, a, i}} - V^{*, \gamma}_{M(s, a)}\right)\right)\leq 3\beta_t,
    \label{eq: tosee}
\end{align}
and then the first inequality in (i) follows by induction and the second inquality in (i) follows naturally. In fact, to see \eqref{eq: tosee}, first note that by Lemma~\ref{lem: basics}.(i) we have that 
\begin{align*}
    \beta_t\leq \sum_{i = 1}^{t} a_t^{i}b_i\leq 2\beta_t 
\end{align*}
and the previous arguments showed that with probability at least $1 - \frac{p}{3}$ we have that for any $s, a, t$, 
\begin{align*}
\left\lvert\sum_{i = 1}^{t}\alpha_{t}^i\left(r_{s, a, i} - \bar{R}(s, a) + \gamma\left(V^{*, \gamma}_{s'_{s, a, i}} - V^{*, \gamma}_{M(s, a)}\right)\right)\right\rvert \leq\beta_t. 
\end{align*}
This concludes the proof that (i) is true with probability at least $1 - \frac{p}{3}$.

Next we focus on (ii). Let $y_h = \bar{r}_{h} - r_h  + \gamma\left(V^{*, \gamma}_{M(s_h, a_h)} - V^{*, \gamma}_{s_{h + 1}}\right)$. We can see that $\left\{y_h\right\}_{i = 0}^{T - 1}$ is a Martingale difference sequence and $\lvert y_h\rvert \leq  \frac{1}{1 - \gamma}$, therefore by Azuma-Hoeffding inequality we have that with probability at least $1 - \frac{2p}{3}$,
\begin{align*}
    \sum_{h = 1}^{T - 1} y_h
    \leq\frac{1}{1 - \gamma} \sqrt{2T\ln\left(\frac{3}{2p}\right)} = \xi_T.
\end{align*}
This concludes the proof that (ii) is true with probability at least $1 - \frac{2p}{3}$.
\end{document}